%% file: neurips_2022.tex
\documentclass{article}

\PassOptionsToPackage{numbers, compress}{natbib}
\usepackage[preprint]{neurips_2022}

\usepackage[utf8]{inputenc} % allow utf-8 input
\usepackage[T1]{fontenc}    % use 8-bit T1 fonts
\usepackage[pagebackref,breaklinks,colorlinks]{hyperref}      % hyperlinks
\usepackage{url}            % simple URL typesetting
\usepackage{booktabs}       % professional-quality tables
\usepackage{amsfonts}       % blackboard math symbols
\usepackage{nicefrac}       % compact symbols for 1/2, etc.
\usepackage{microtype}      % microtypography
\usepackage{xcolor}         % colors
\usepackage{soul}
\usepackage{amsmath}
\usepackage{mathtools}
\usepackage{graphicx}
\usepackage{xspace}
\usepackage{array}
\usepackage{bm}
\usepackage{caption}
\usepackage{multirow}
\usepackage{wrapfig}
\usepackage{algorithm}
\usepackage{listings}
\usepackage{eucal}
\usepackage{booktabs,subcaption,amsfonts,dcolumn}
\usepackage{color}
\usepackage{colortbl}
\usepackage{arydshln}

\definecolor{airforceblue}{rgb}{0.36, 0.60, 0.84}
\definecolor{palegreen}{rgb}{0.66, 0.82, 0.55}

\newlength\savewidth\newcommand\shline{\noalign{\global\savewidth\arrayrulewidth
  \global\arrayrulewidth 1pt}\hline\noalign{\global\arrayrulewidth\savewidth}}
\newcommand{\tablestyle}[2]{\setlength{\tabcolsep}{#1}\renewcommand{\arraystretch}{#2}\centering\footnotesize}
\renewcommand{\paragraph}[1]{\vspace{1.25mm}\noindent\textbf{#1}}

\newcolumntype{x}[1]{>{\centering\arraybackslash}p{#1pt}}
\newcolumntype{y}[1]{>{\raggedright\arraybackslash}p{#1pt}}
\newcolumntype{z}[1]{>{\raggedleft\arraybackslash}p{#1pt}}

\newcommand{\app}{\raise.17ex\hbox{$\scriptstyle\sim$}}

\definecolor{deemph}{gray}{0.6}
\newcommand{\gc}[1]{\textcolor{deemph}{#1}}
\definecolor{baselinecolor}{gray}{.9}
\newcommand{\baseline}[1]{\cellcolor{baselinecolor}{#1}}

\title{Video-based Human-Object Interaction Detection from Tubelet Tokens}
\author{%
            Danyang Tu$^1$, 
            Wei Sun $^1$,
	        Xiongkuo Min$^1$, 
	        Guangtao Zhai$^{1}$,
	        Wei Shen$^2$  \\
	        $^1$Institute of Image Communication and Network Engineering, Shanghai Jiao Tong University\\
		$^2$MoE Key Lab of Artificial Intelligence, AI Institute, Shanghai Jiao Tong University\\
		}
\begin{document}

\maketitle

\begin{abstract}
    We present a novel vision Transforme\underline{\textbf{r}}, named TUTOR, which is able to learn \underline{\textbf{tu}}belet \underline{\textbf{to}}kens, served as highly-abstracted spatiotemporal representations, for video-based human-object interaction (V-HOI) detection. The tubelet tokens structurize videos by agglomerating and linking semantically-related patch tokens along spatial and temporal domains, which enjoy two benefits: 1) Compactness: each tubelet token is learned by a selective attention mechanism to reduce redundant spatial dependencies from others; 2) Expressiveness: each tubelet token is enabled to align with a semantic instance, \emph{i.e.}, an object or a human, across frames, thanks to agglomeration and linking. The effectiveness and efficiency of TUTOR are verified by extensive experiments. Results shows our method outperforms existing works by large margins, with a relative mAP gain of $16.14\%$ on VidHOI and a 2 points gain on CAD-120 as well as a $4 \times$ speedup.
\end{abstract}

\section{Introduction}

\begin{figure}[t]
\centering
\includegraphics[width=\linewidth]{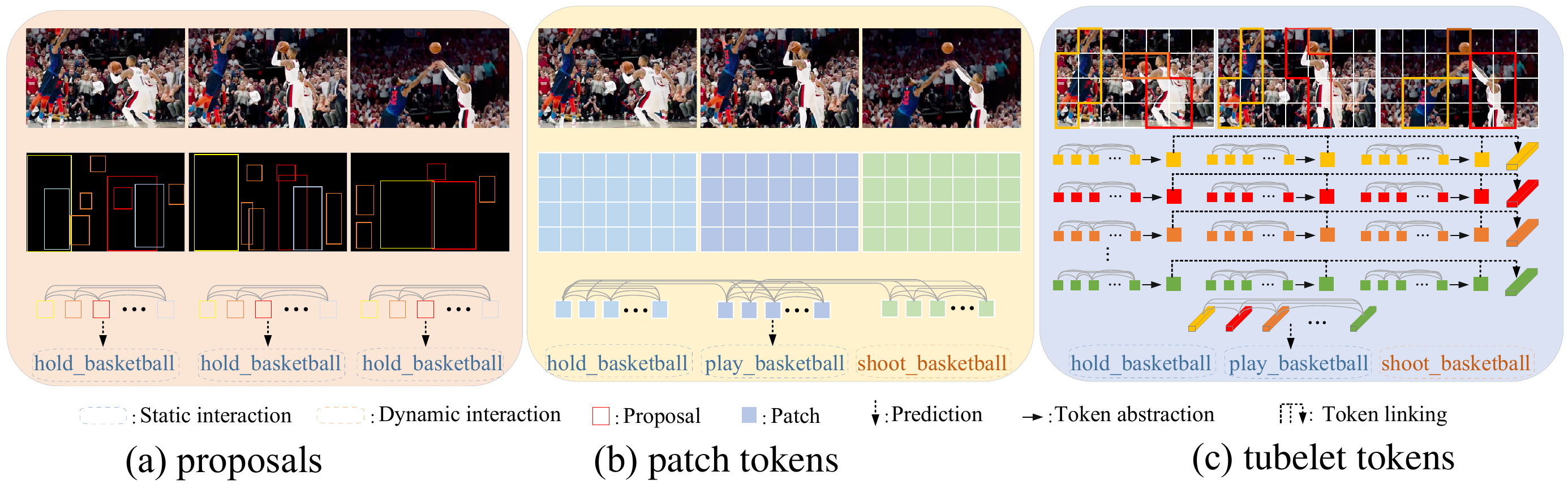}
\caption{Illustration of different strategies for V-HOI detection, which are built on different representations. (a) Image-based HOI detection methods, which process each frame as \emph{i.i.d} data, and recognize the interrelations among pre-detected proposals in each frame independently. (b) ViT-like V-HOI detection frameworks, which perform global attention mechanism on patch tokens over space and time. (c) Our proposed TUTOR, which structurizes a video into a few tubelet tokens by token abstraction along spatial domain and token linking along temporal domain.} 
\label{first_img}
\end{figure}

    Human-object interaction (HOI) detection is a detailed scene understanding task, which requires both localization of interacted human-object pairs and recognition of interaction labels. Existing methods mostly investigated detecting HOIs in static images without capturing temporal information (Figure~\ref{first_img} (a)), thus lack the ability to detect time-related interactions (\emph{e.g.}, \texttt{shoot} or \texttt{pass} a basketball). However, interactions are more of time-related in practical scenario, leading to a strong demand to detect HOIs from videos, a more challenging problem built on spatiotemporal semantic representations. 
    
    Transformer, originated from natural language processing (NLP), is an intuitive choice for its eminent capability of reasoning long-range dependencies, in which one of the most crucial components is the token. A token serves as an element of data representations, which is usually a word in language. However, unlike language that naturally has such a discrete signal space for building tokenized dictionaries, images lie in a continuous and high-dimensional space. To address this issue, vision Transformer (ViT)~\cite{dosovitskiy2020image} provided a solution that divides each image into several local patch tokens (``visual words'') to structurize the entire image as a ``visual sentence''(Figure~\ref{first_img}(b)). This solution has become a \emph{de facto} tokenization standard followed by most existing Transformer-based methods, which has achieved excellent performance for various vision tasks, especially image classification.
    
    Nevertheless, this patch based tokenization strategy might not be proper for video-based HOI (V-HOI) detection (as the performance degradation of the ViT-like framework shown in Table.~\ref{tab:spatial}). We find that the reason is the patch tokens generated by regular splitting are difficult to exactly capture instance-level semantics (an instance is an object or a human, \emph{e.g.}, the basketball shooter in Figure~\ref{first_img}), which yet is crucial for V-HOI detection to reason the interaction labels. These patch tokens inevitably suffer from redundancy due to an information mixture from different instances as well as insufficiency due to only a part occupancy of an instance, which limit their representation ability.
    
    In this paper, we present TUTOR, a new Transforme\textbf{R} for V-HOI detection built on \textbf{TU}belet \textbf{TO}kens, handling aforementioned limitations favorably. The tokenization of the tubelet tokens is not based on fixed regular splitting but is jointly performed with the learning of the Transformer encoder. This enables the tubelet tokens to progressively emerge and represent high-level visual semantics. Concretely, first, along the spatial domain, we alternatively update the representation for each patch token by a selective attention mechanism and agglomerate semantically-related patch tokens into instance tokens. The selective attention mechanism ensures that attention is performed among tokens expected to belong to the same instance, which reduce redundant spatial dependencies from others. Then, along the temporal domain, we link instance tokens across frames to form the tubelet tokens. Figure~\ref{first_img}(c) illustrates the process of tubelet token generation.  Experimental results show that TUTOR outperforms existing sota methods by large margins. Specifically, we achieve a relative mAP gain of $16.4\%$ on VidHOI~\cite{chiou2021st} and a 2 points F1 score gain CAD-120~\cite{koppula2013learning}, with a $4 \times$ inference speedup.

%\vspace{-15pt}
\section{Related Work}
%\vspace{-13pt}
{\bfseries HOI detection.} 
    HOI detection has attracted considerable research interests recently for its great potential in high-level visual understanding tasks, such as captioning~\cite{HIC}, visual question answering~\cite{garcia2020knowledge}. Most previous works are devoted to detecting HOIs in static images~\cite{chao2018learning,drg,ican,interactions,no-frills,zhi_eccv2020,uniondet,kim2020detecting,li2020detailed,li2019transferable,lin2020action,liu2020amplifying,qi2018learning,ulutan2020vsgnet,wan2019pose,wang2020contextual,wang2019deep,xu2019interact,yang2020graph,zhong2021polysemy,zhou2019relation,zhou2020cascaded,qpic,endd,hotr,chen2021reformulating,zhang2021mining}. Without considering temporal information, these methods fail to detect time-related interactions, restricting their value in practical applications. In contrast, video-based HOI detection is a more practical problem, which however is less explored~\cite{qi2018learning,nagarajan2019grounded,nawhal2020generating,sunkesula2020lighten,chiou2021st,wang2021spatio,ji2021detecting}. \cite{qi2018learning,sunkesula2020lighten, wang2021spatio} detected HOIs in videos by building graph neural networks to capture spatiotemporal information. In~\cite{nagarajan2019grounded}, HOI ``hotspots'' can be directly learned from videos by jointly training a video-based action recognition network as well as an anticipation model. Inspired by image-based methods, \cite{chiou2021st} introduced a two-stage framework where the frame-wise human/object features are firstly extracted by using trajectories, and then HOIs are detected by processing the instance features as well as auxiliary features, including spatial configurations and human poses. However, these methods lack the ability to model long range contextural information, resulting in poor performance when the interacted human and object are far apart. Inspired by the strong ability of Transformer in modeling long-range dependencies, \cite{ji2021detecting} proposed a spatiotemporal Transformer to reason human-object relationships in videos, which detects human/object proposals firstly and then captures spatial and temporal information by using two dense-connected Transformers, respectively. However, such dense-connected manner introduces extra computation and ambiguity in token representation as mentioned above.\\
{\bfseries Transformer in video analysis.} 
    Transformer~\cite{attention} originated from natural language processing (NLP), and has been widely explored in vision tasks recently. Specifically, Transformer has also show a great potential in video analysis, \emph{e.g.}, action recognition~\cite{liu2021end, zhang2022actionformer}, video restoration~\cite{liang2022vrt}, video question answering~\cite{garcia2020knowledge}, video instance segmentation~\cite{wang2021end} and etc. However, most spatiotemporal Transformer follow the \emph{de facto} scheme of ViT~\cite{dosovitskiy2020image}, \emph{i.e.}, simply dividing an image into local patches and stacking global attention, which lacks sufficient exploration of the properties of visual signal, thus suffering from insufficiency token representation and explosive computation. 

%\vspace{-15pt}
\section{Methodology}
\begin{figure}[t]
    %\vspace{-0.3cm}
    \centering
    \includegraphics[width=\linewidth]{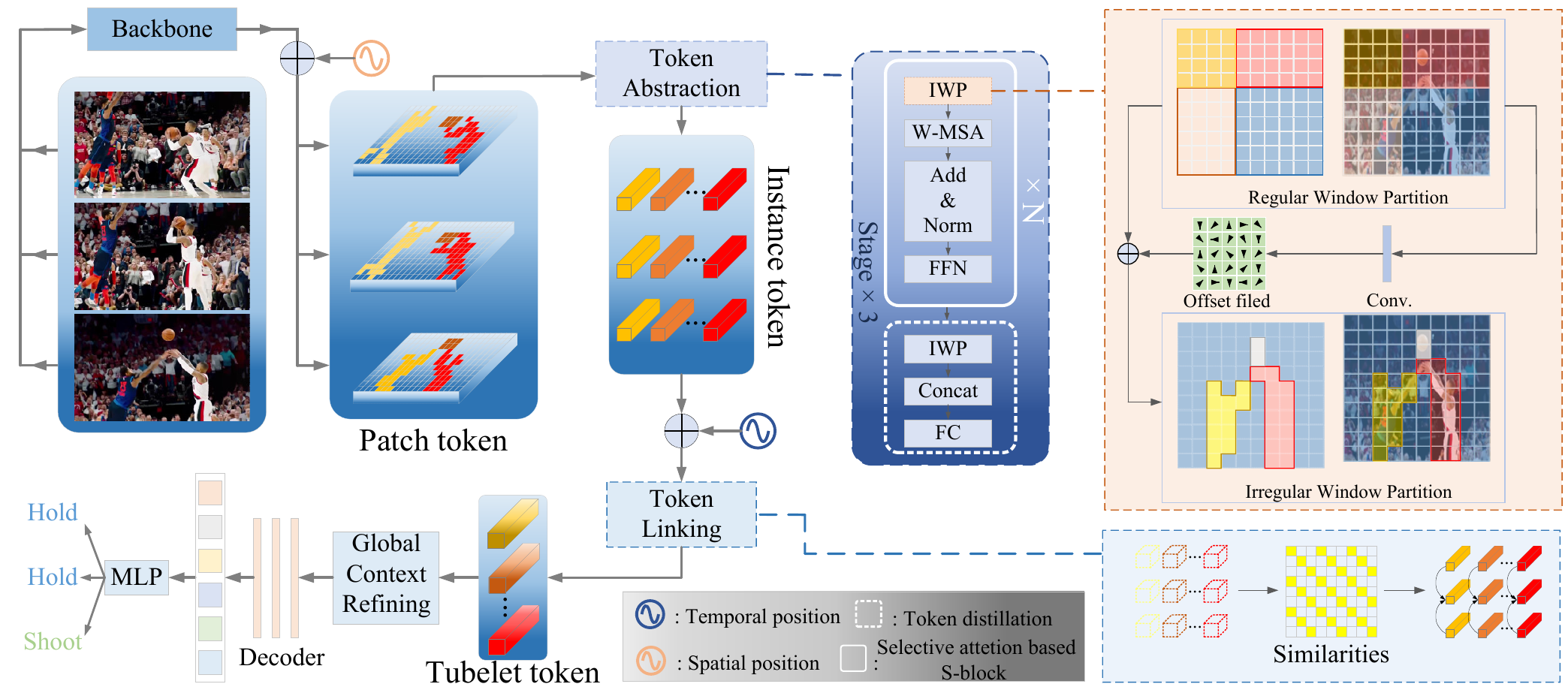}
    %\captionsetup{size=small}
    \caption{{\bfseries The architecture of TUTOR.} It consists of 1) a backbone to generate the initial patch tokens; 2) a token abstraction module that alternatively update token representations and agglomerate patch tokens, to progressively form instance tokens; 3) a token linking module that links the semantically related instance tokens across different frames to form tubelet tokens; 4) a simple global attention layer to reinforce the global contextual information and 5) a standard Transformer decoder to decode the HOI instances. We use different dashed squares to zoom in on different key modules.}
    %\vspace{-17pt}
    \label{whole_model}
\end{figure}

%\vspace{-11pt}
 The main idea of TUTOR is to structurize a video into a few tubelet tokens, which serve as highly-abstracted spatiotemporal representations. Towards this end, we propose a reinforced tokenization strategy, which jointly performs tokenization and optimization of the Transformer encoder, as illustrated in Figure~\ref{whole_model}. The process of tubelet token generation consists of two steps: 1) Token abstraction along the spatial domain, where patch tokens are alternatively updated by a selective attention mechanism and agglomerated into instance tokens; 2) Token linking along the temporal domain, where instance tokens across frames are linked to form tubelet tokens. In this section, we describe these two steps in details.

%\vspace{-11pt}
\subsection{Backbone}
%\vspace{-9pt}

    Taking a video clip $\mathbf{x} \in \mathbb{R}^{T \times H \times W \times 3}$ that consists of $T$ frames with size $H \times W$ as the input, we use a ResNet~\cite{he2016deep} followed by a feature pyramid network (FPN)~\cite{lin2017feature} as the backbone on $t$-th frame to generate a feature map  $\mathbf{z}_{(t,b)}\in \mathbb{R}^{\frac{H}{4} \times \frac{W}{4} \times C_0}$, where $t=1,2,..,T$ and $C_0=32$ is the channel number of the initial feature map. 

%\vspace{-13pt}
\subsection{Token Abstraction}
%\vspace{-9pt}
    %Token agglomeration, working on spatial domain, is organized in $3$ \emph{Stage}s through a hierarchy of Transformer layers, where each stage contains multiple selective attention mechanism to perform token representation learning, and an agglomeration layer that merges and transforms patch tokens into corresponding instance tokens.\\
    Token abstraction along the spatial domain is organized in $3$ \emph{Stage}s through a hierarchy of Transformer layers. Each stage performs token representation learning by a selective attention mechanism and merges semantically-related patch tokens into instance tokens by an agglomeration layer.  Here, we denote the feature map of $t$-th frame inputted into $s$-th stage  as $\mathbf{z}_{(t,s)} \in \mathbb{R}^{H_s \times W_s \times C_s}$. Specifically, $\mathbf{z}_{(t,1)} = \mathbf{z}_{(t,b)}$.\\
{\bfseries Selective attention.}
 To eliminate the redundancy caused by information mixture from different instances,  we are motivated to selectively calculate attention weights among related tokens, \emph{i.e.}, tokens belong to the same instances. To this end, we propose an irregular window partition (IWP) mechanism ({\color{orange}{orange}} rectangle in the right of Figure~\ref{whole_model}), a simple yet effective strategy that samples a group of related tokens into a local window. IWP is inspired from regular window partition (\emph{e.g.}, Swin~\cite{swin}), where the tokens are grouped by sliding a regular rectangle $\mathcal{R}$ with size of $S_w \times S_w$ over the feature map $\mathbf{z}_{(t,s)}$ in $s$-th stage. For instance,
    $\mathcal{R} =  \{[0,0], [0, 1],...,[3,4], [4, 4]\}$
    defines a regular window with size of $5 \times 5$. Then, for the $i$-th regular window, we have
    \begin{equation}
        \mathcal{Z}_{\text{rw}}^i =\{\mathbf{z}_{(t,s)}(\mathbf{p}_n + [(i-1)S_w, (i-1)S_w])\;|\;\mathbf{p}_n \in \mathcal{R}\},
        \label{regular_window}
    \end{equation}
    where $\mathbf{z}_{(t,s)}([i,j]) \in \mathcal{R}^{1 \times C_s}$ denotes the feature vector at spatial location $[i,j]$, which is also the representation of $(i{\cdot}j)$-th token, unless specified below. However, as shown in Figure~\ref{whole_model}, regular windows can easily divide an instance into several parts due to the limitation of a fixed shape, leads to unrelated tokens within a window, \emph{i.e.}, belonging to different instances. IWP makes a simple change by augmenting regular grid $\mathcal{R}$ with learned offsets, which allows the generated irregular windows to be aligned with humans/objects with arbitrary shapes. With offsets $\{\Delta\mathbf{p}_n|n=1,2,...,N\}$ and $N=|\mathcal{R}|$, for the tokens in $i$-th irregular window, we have
    \begin{equation}
        \mathcal{Z}_{\text{irw}}^i =\{\mathbf{z}_{(t,s)}(\mathbf{p}_n + [(i-1)S_w, (i-1)S_w] + \Delta\mathbf{p}_n) \;|\;\mathbf{p}_n \in \mathcal{R}\}.
        \label{irregular_window}
    \end{equation}
    Specifically, $\Delta\mathbf{p}_n$ are learned by performing a convolutional layer with kernel size of $3 \times 3$ over the input feature map $\mathbf{z}_{(t,s)}$. As the offsets are typically fractional, the right part of Eq.~\ref{irregular_window} is implemented practically via bilinear interpolation as
    \begin{equation}
        \mathbf{z}_{(t,s)}(\mathbf{p}) = \sum_{\mathbf{q}}\text{B}(\mathbf{q},\mathbf{p})\cdot\mathbf{z}_{(t,s)}(\mathbf{q}),
    \end{equation}
    where $\mathbf{p}=\mathbf{p}_n + [(i-1)S_w, (i-1)S_w] + \Delta\mathbf{p}_n$ denotes an arbitrary location, $\mathbf{q}$ enumerates all integral locations, and $\text{B}$ is the bilinear interpolation kernel.\\
    On this basis, we alternatively update the token representation by stacking several \emph{S-blocks} (rectangle in white solid line in Figure~\ref{whole_model})  that are built on selective attention, \emph{i.e.}, performing attention mechanism within irregular windows. Specifically, each block is computed as
    \begin{align}
        &{\hat{\mathbf{z}}_{\text{irw}}^l} = \text{IWP}(\mathbf{z}^{l-1}, S_w),  \\
        &{\hat{\mathbf{z}}^l} = \text{W-MSA}(\text{LN}(\hat{\mathbf{z}}_{\text{irw}}^l + \mathbf{p}_s^l)) + \text{Flatten}(\mathbf{z}^{l-1}), \\
        &{\mathbf{z}^l} = \text{Reshape}(\text{MLP}(\text{LN}(\hat{\mathbf{z}}^l)) + \hat{\mathbf{z}}^l),
    \end{align}
    where $\mathbf{z}^l$ is updated representation for all tokens, $\mathbf{p}_s^l$ is sine-based spatial position encoding at $l$-th \emph{S-block}, and $\hat{\mathbf{z}}$ denotes various intermediate variables. The detailed calculation process can be found in \emph{supplementary file}. Here, we factorize the conventional 3D position encoding into a 2D spatial position encoding and a 1D temporal one since the spatial and temporal information are separately extracted. In detail, ``$\text{W-MSA}$'' denotes window-based multi-head self-attention, ``$\text{LN}$'' is layer normalization and ``$\text{MLP}$'' refers to multi-layer perceptron. Since the convolution layer in IWP is operated on 2D feature map yet attention is calculated on sequential features, we use a ``$\text{Flatten}$'' ($2D \rightarrow 1D$) operation to collapse the spatial dimension as well as a ``$\text{Reshape}$'' ($1D \rightarrow 2D$) operation to restore it. The computational complexity of a global MSA (G-MSA) block and a irregular-window-based block (IW-MSA) for total $T$ frames at $s$-th stage are respectively:
    \begin{align}
        &{\Omega(\text{G-MSA})} = 4H_sW_sTC_s^2 + 2(H_sW_sT)^2C_s,\\
        &{\Omega(\text{IW-MSA})} = 4H_sW_sTC_i^2 + 2(S_w^2+K^2)H_sW_sTC_s,
    \end{align}
    where $K=3$ is the kernel size of convolutional layer and $S_w$ is fixed as $7$. In comparison, IW-MSA can effectively reduce the computational complexity.  In our experiment, the number of \emph{S-block} for (1-3)-th stage is set to 1, 1, 3, respectively.\\
{\bfseries Token agglomeration.} 
     We perform token agglomeration at the end of each stage to merge semantically similar tokens. Specifically, we first perform IWP with a window size of $2 \times 2$ to dynamically sample every $4$ related tokens into a window. Then, we concatenate the tokens within each window and apply a fully-connected (FC) layer on the concatenated $4C_s$-dimensional features (each token is represented by a feature vector with $C_s$ channels at $s$-th stage). We set the output dimension to $2C_s$. This process reduces the number of  tokens by a multiple of $2 \times 2 = 4$ after each stage.
     
    To sum up, token abstraction totally reduces the number of tokens by a factor of $4^3=64$ and increases the dimension by a factor of $2^3=8$. It structures each frame into a few instance tokens on the basis of selective attention and token agglomeration, which reduces the visual redundancy and also enjoys the advantage of Transformer with a affordable computational costs.
%\vspace{-11pt}
\subsection{Token Linking}
%\vspace{-9pt}

   Token linking aims at forming compact and expressive tubelet tokens by linking instance tokens with the same semantic along the temporal domain. Specifically, assuming that after token agglomeration, each frame is structured as $N$ instance tokens. Then a video clip of $T$ frames can be denoted as $\mathcal{Z}_{\text{ins}} = \{\mathbf{z}_t^i | i=1,2,...,N;t=1,2,...,T\}$, where $\mathbf{z}_t^i$ refers to the $i$-th token in the $t$-th frame. Here, a sine-based 1D temporal position encoding is additionally added to $\mathcal{Z}_{\text{ins}}$. The goal of token linking is to link $T$ instance tokens with the same semantic across $T$ frames (one per frame), so that the video clip can be structurized as $N$ spatiotemporal tubelet tokens. To this end, we propose an exemplar based between-frame one-to-one matching strategy. We first chose $\mathcal{Z}_q = \{\mathbf{z}_q^i  | i=1,2,...,N\}$ as the exemplar frame, where $q = \left \lfloor  \frac{T}{2} \right \rfloor$ is the index of the middle frame in the video clip. We denote the tokens in the exemplar frame and those in rest frames $\mathcal{Z}_k = \{\mathbf{z}_t^i | \mathbf{z}_t^i \in \mathcal{Z}_{\text{ins}}, t \neq r \}$ as \emph{query} tokens and \emph{key} tokens, respectively. Then, we compute a similarity matrix $\mathbf{A}$ between the query tokens and the key tokens via a \texttt{Gumbel-Softmax}~\cite{jang2016categorical} operation computed over the query tokens as
    \begin{equation}
        \mathbf{A}_t^{(i,j)} = \frac{\exp(W_q\mathbf{z}_q^i \cdot W_k\mathbf{z}_t^j + \gamma_i)}{\sum_{n=1}^{N}\exp(W_q\mathbf{z}_q^n \cdot W_k\mathbf{z}_t^j + \gamma_n)} \;\; s.t. \;\; t\neq r,
        \label{eq:softmax}
    \end{equation}
    where $W_q$ and $W_k$ are the weights of the learned linear projections for the query tokens and the key tokens, respectively, and $\gamma_i$s are \emph{i.i.d} random samples drawn from the \texttt{Gumbel(0,1)} distribution that enables the \texttt{Gumbel-Softmax} distribution to be close with the real categorical distribution. 
    Then, we introduce a modified \texttt{nms-one-hot} operation to determine the one-to-one correspondence between the query tokens and the key tokens of each frame. Specifically, for $n$-th key token $\mathbf{z}_t^n$ in $t$-th frame, the norm \texttt{one-hot} assignment is performed by taking the value of $\texttt{argmax}\{\mathbf{A}_t^{(i,n)}|i=1,2,...,N\}$. However, such an operation cannot ensure a one-to-one correspondence, \emph{i.e.}, more than one key tokens in a frame could be assigned to the same query token. To address this issue, in \texttt{nms-one-hot} scheme, for example, when $n$-th and $m$-th key token in $t$-th frame are assigned to $i$-th query token simultaneously, we assign the the one with a higher similarity, \emph{i.e.}, $\texttt{max}(\mathbf{A}_t^{(i,n)}, \mathbf{A}_t^{(i,m)})$, to the $i$-th query token. Then, if $\mathbf{z}_t^n$ has been assigned to the $i$-th query token, we manually set $\mathbf{A}_t^{(i,m)}$ as $0$ and continue to conduct the assignment operation. Since the \texttt{nms-one-hot} operation is not differentiable, we adopt the straight through strategy in~\cite{esser2021taming} to compute the assignment matrix:
    \begin{equation}
        \hat{\mathbf{A}} = \texttt{nms-one-hot}(\mathbf{A}) + \mathbf{A} - \texttt{sg}(\mathbf{A}),
    \end{equation}
    where $\texttt{sg}(\cdot)$ is the stop gradient operator. $\hat{\mathbf{A}}$ is numerically equal to \texttt{nms-one-hot} assignments and its gradient is equal to the gradient of $\mathbf{A}$, which makes the token linking module differentiable and end-to-end trainable. Finally, we link the tokens corresponding to the same query token to form the tubelet tokens $\mathcal{Z}_{\text{tube}} = \{\mathbf{z}_{\text{tube}}^i|i=1,2,...,N\}$, which is computed as
    \begin{equation}
        \mathbf{z}_{\text{tube}}^i = \mathbf{z}_r^i + W_o \frac{\sum_{t=1}^{T}\hat{\mathbf{A}}_t^{(i,\phi(i,t))}W_v\mathbf{z}_t^{\phi(i,t)}}{\sum_{t=1}^{T}\hat{\mathbf{A}}_t^{(i,\phi(i,t))}} \;\;\; s.t. \;\;\; t \neq r,
        \label{eq:temporal}
    \end{equation}
    where $W_o$ and $W_v$ are the learned weights of projectors, and $\phi(i,t)$ is the index of token in $t$-th frame and being assigned to $i$-th query token.

%\vspace{-15pt}
\subsection{Global Context Refining}
%\vspace{-11pt}
    After token agglomeration and linking, a spatiotemporal video representation is structurized as a few tubelet tokens. On this basis, we perform an additional global attention layer to model global contextual information. Our intuition is two-fold: 1) Global context can significantly boost the performance of interaction recognition, \emph{e.g.}, if grassland is detected, a person is more likely to be playing soccer than basketball. 2) Different interactions can be co-occurring, \emph{e.g.}, a person is \texttt{holding} a fork could be \texttt{eating} something.

%\vspace{-15pt}
\subsection{Decoder \& Prediction Head}
%\vspace{-11pt}

{\bfseries Decoder.} 
    The decoder follows the standard architecture of the Transformer, which transforms $N_q$ embeddings by stacking $6$ layers consisting of self-attention and cross-attention mechanisms. These embeddings are learned position encodings which are initialized to constants and we refer them to as HOI \emph{queries}. Being added to the input of each attention layer, the $N_q$ queries are transformed as output embeddings by the decoder, which performs global reasoning by using the entire video clip as context.\\
{\bfseries Prediction head.}
    Following~\cite{qpic}, the prediction head is composed of four feed-forward networks (FFNs): human-bounding-box FFNs$f_h$, object-bounding-box $f_o$, object-class FFNs$f_o^c$, and action-class FFNs $f_{a}^c$. Specifically, $f_h$ and $f_o$ are both a 3-layer perceptron followed by a sigmoid function, which output normalized human- and object-bounding box $\hat{\mathbf{b}}_h \in [0,1]^4$, $\hat{\mathbf{b}}_o \in [0,1]^4$, respectively. $f_o^c$ is a linear layer followed by a softmax function, predicting the probability of object classes $\hat{\mathbf{c}}_o \in [0,1]^{N_{obj}+1}$, where $N_{obj}$ is the number of object classes and the $(N_{obj}+1)$-th element in $\hat{\mathbf{c}}_o$ indicates the query has no corresponding human-object pair. Since actions could be co-occurring, $f_a^c$ is a linear layer followed by a sigmoid function rather than the softmax function. It outputs the probability of action classes $\hat{\mathbf{c}}_a \in [0,1]^{N_{act}}$, which has no an additional element to indicate \emph{no-action}. Here, $N_{act}$ is the number of action classes.
    
%\vspace{-11pt}
\subsection{Loss Function}
%\vspace{-7pt}

    We follow the loss calculation scheme in~\cite{qpic}, including bipartite matching and loss calculation.\\
{\bfseries Bipartite matching.} 
    The model outputs a fixed-size set of $N_q$ predictions, which is denoted as $\mathcal{O} = \{o^i\}_{i=1}^{N_q}$. We use the $\mathcal{G} = \{g^i|i=1,2,...,M, \emptyset_1, \emptyset_2,...,\emptyset_{N_q-M}\}$ to represent the padded groundtruths, where $M$ is the real number of HOI instances in a video clip and $\emptyset$ refers to \emph{no-instance}. Then, the matching process can be formulated as an injective function: $\omega_{T \rightarrow O}$, being computed as
    \begin{equation}
        \mathcal{L}_{cost} = {\textstyle \sum_{i=1}^{N_q}}\mathcal{L}_{match}(g^i,o^{\omega(i)}),
    \end{equation}
    where $\omega(i)$ is the index of predicted HOI instance assigned the $i$-th groundtruth, and $\mathcal{L}_{match}(t^i,o^{\omega(i)})$ is the matching cost between the $i$-th groundtruth and $(\omega(i))$-th prediction, being calculated as
    \begin{equation}
        \mathcal{L}_{match}(g^i,o^{\omega(i)}) = \alpha_1\mathcal{L}_{b}^i + \alpha_2\mathcal{L}_o^i + \alpha_3\mathcal{L}_a^i.
        \label{cost_function}
    \end{equation}
     $\mathcal{L}_b$ is regression cost between groundtruths $\mathbf{b}$ and predictions $\hat{\mathbf{b}}$ for human and object boxes:
    \begin{equation}
        \mathcal{L}_b^i = \beta_1[\|\mathbf{b}_h^i - \hat{\mathbf{b}}_h^{\omega(i)} \| + \|\mathbf{b}_o^i - \hat{\mathbf{b}}_o^{\omega(i)} \|] - \beta_2[\text{GIoU}(\mathbf{b}_h^i, \hat{\mathbf{b}}_h^{\omega(i)}) + \text{GIoU}(\mathbf{b}_o^i, \hat{\mathbf{b}}_o^{\omega(i)})];
        \label{box_loss}
    \end{equation}
    $\mathcal{L}_o$ is recognition cost between groundtruths $\mathbf{c}_o$ (one-hot vector) and predictions $\hat{\mathbf{c}}_o$ for objects:
    \begin{equation}
        \mathcal{L}_o^i = -\hat{\mathbf{c}}_o({\omega(i)}) \;\; s.t. \;\; \mathbf{c}_o(i) = 1;
    \end{equation}
    $\mathcal{L}_a$ is recognition cost between groundtruths $\mathbf{c}_a$ and predictions $\hat{\mathbf{c}}_a$ for actions:
    \begin{equation}
        \mathcal{L}_a^i = -\frac{1}{2}\left(\frac{{\mathbf{c}_a^i}^{\top}\hat{\mathbf{c}}_a^{\omega(i)}}{\|\mathbf{c}_a^i\|_1 + \epsilon} + \frac{(1-\mathbf{c}_a^i)^{\top}(1-\mathbf{c}_a^{\omega(i)})}{\|1-\mathbf{c}_a^i\|_1 + \epsilon} \right),
    \end{equation}
    which takes both positive and negative action classes into account by using the weighted average of the two with the inverse number of nonzero elements as weight rather than the vanilla one. Finally, we follow DETR~\cite{deformableDETR} to use Hungarian algorithm~\cite{kuhn1955hungarian} to determine the optimal assignment $\hat{\omega}$ among the set of all possible permutations of $N_q$ elements $\mathbf{\Omega}_{N_q}$, which is formulated as
    \begin{equation}
        \hat{\omega} = \underset{\omega \in \mathbf{\Omega_{N_q}}}{\arg \min} \; \mathcal{L}_{cost}.
    \end{equation}
{\bfseries Loss calculation.}
After the optimal one-to-one matching between the groundtruths and the predictions is found, the
loss to be minimized in the training phase is calculated as $\mathcal{L} = \eta_1\mathcal{L}_b + \eta_2\mathcal{L}_o + \eta_3\mathcal{L}_a$, where $\mathcal{L}_b$ is defined as the same as Eq.~\ref{box_loss} and $\mathcal{L}_o$ is the CE loss and $\mathcal{L}_a$ is the focal loss.

%\vspace{-11pt}
\section{Experiments}
%\vspace{-9pt}
     %In this section, we conduct comprehensive experiments to take a closer look at spatiotemporal Transformer for V-HOI detection. Following a briefly introduction of the experimental datasets and the implementation details, we first analysis the advantages of CNN-based and Transformer-based methods, respectively. Then, a detailed experimental comparison of different strategies to capture spatial and temporal information is shown in Sec. Next, we present the ablation study to validate the effectiveness of our method. At last, we compare our method with state-of-the-art methods.
 
 %%\vspace{-11pt}
\subsection{Datasets \& Metrics}
%\vspace{-9pt}
    We conduct experiments on VidHOI~\cite{chiou2021st} and CAD-120~\cite{koppula2013learning} benchmarks to evaluate the proposed methods by following the standard scheme. VidHOI is a large-scale dataset for V-HOI detection, comprising 6,366 videos for training set and 756 videos for validation. In VidHOI, 50 relation categories are annotated, of which half are time-related ones. As same as the widely used image-based HOI detection dataset HICO-DET~\cite{chao2018learning}, mean \texttt{AP} (\texttt{mAP}) is calculated as the evaluation metric for VidHOI, which is reported over three sets: 1)Full: all 557 categories are evaluated; 2)Rare:315 categories with less than 25 instances and 3)Non-rare: 242 categories with more than 25 instances. CAD-120 is a relatively smaller dataset that consists of 120 RGB-D videos. Here, we only use the RGB images and the 2D bounding boxes annotations of humans and objects. Following standard scheme, we calculate the sub-activity \texttt{F1} score as metrics.

%\vspace{-11pt}
\subsection{Implementation Details}
%\vspace{-9pt}
    The dimension of HOI \emph{query} is set to 256, which is the same as the tubelet tokens ($32 \times 2^3$). The number of queries is set to 100 for VidHOI and 50 for CAD-120. Following~\cite{qpic}, we set $\alpha_1$, $\alpha_2$, $\alpha_3$ to 1, 1, 1, and $\beta_1$, $\beta_2$ to 2.5, 1, respectively. For loss calculation, $\eta_{(1,2,3)}$ are all set to 1. To save computational resources, the backbone is initialized by the backbone weights of QPIC~\cite{qpic}, and then frozen without being updated. We employed a AdamW~\cite{adamw} optimizer for 150 epochs. A batch size of $16$ on $8$ RTX-2080Ti GPUs, and learning rate $l_r=2.5e^{-4}$ for Transformer and $1e^{-5}$ for  FPN are used. The $l_r$ decayed by half at 50-th, 90-th and 120-th epoch, respectively. We use a $l_r=10^{-6}$ to warm up the training for the first 5 epochs, and then go back to $2.5e^{-4}$ and continue training.
    
%\vspace{-9pt}
\subsection{Analysis of CNN-based \& Transformer-based Methods}
\begin{figure}[t]
    \centering
    \subfloat[\textbf{Spatial distance}.
    \label{fig:dis}
    ]{
    \begin{minipage}[t]{0.32\textwidth}
    \centering
    \includegraphics[width=0.98\linewidth]{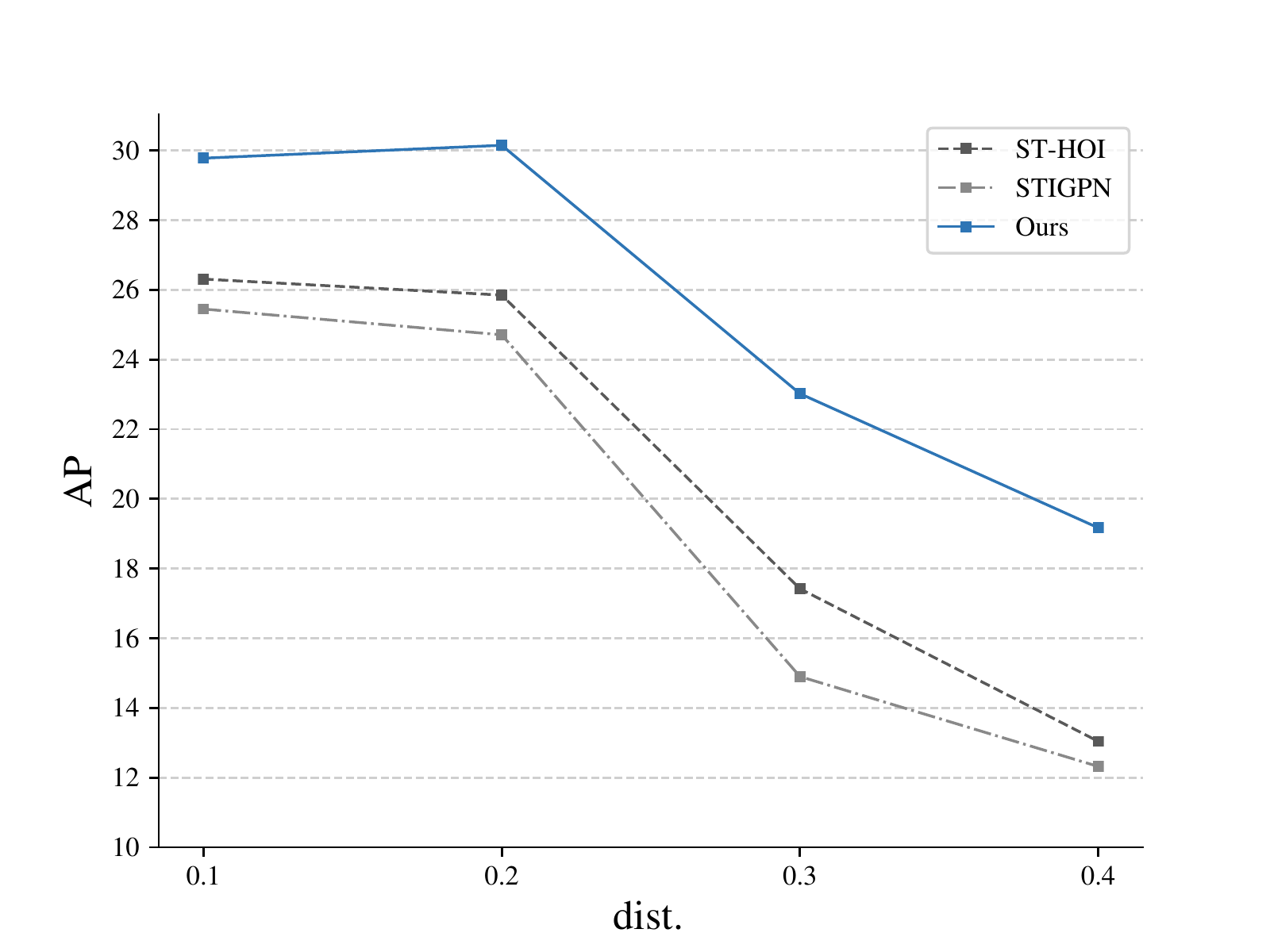}
    %\captionsetup{width=0.95\linewidth}
    %\captionsetup{size=small}
    %\caption{ (a)}
    \end{minipage}
    }
    \subfloat[\textbf{Temporal pooling}.
    \label{fig:tem}
    ]{
    \begin{minipage}[t]{0.32\textwidth}
    \centering
    \includegraphics[width=0.98\linewidth]{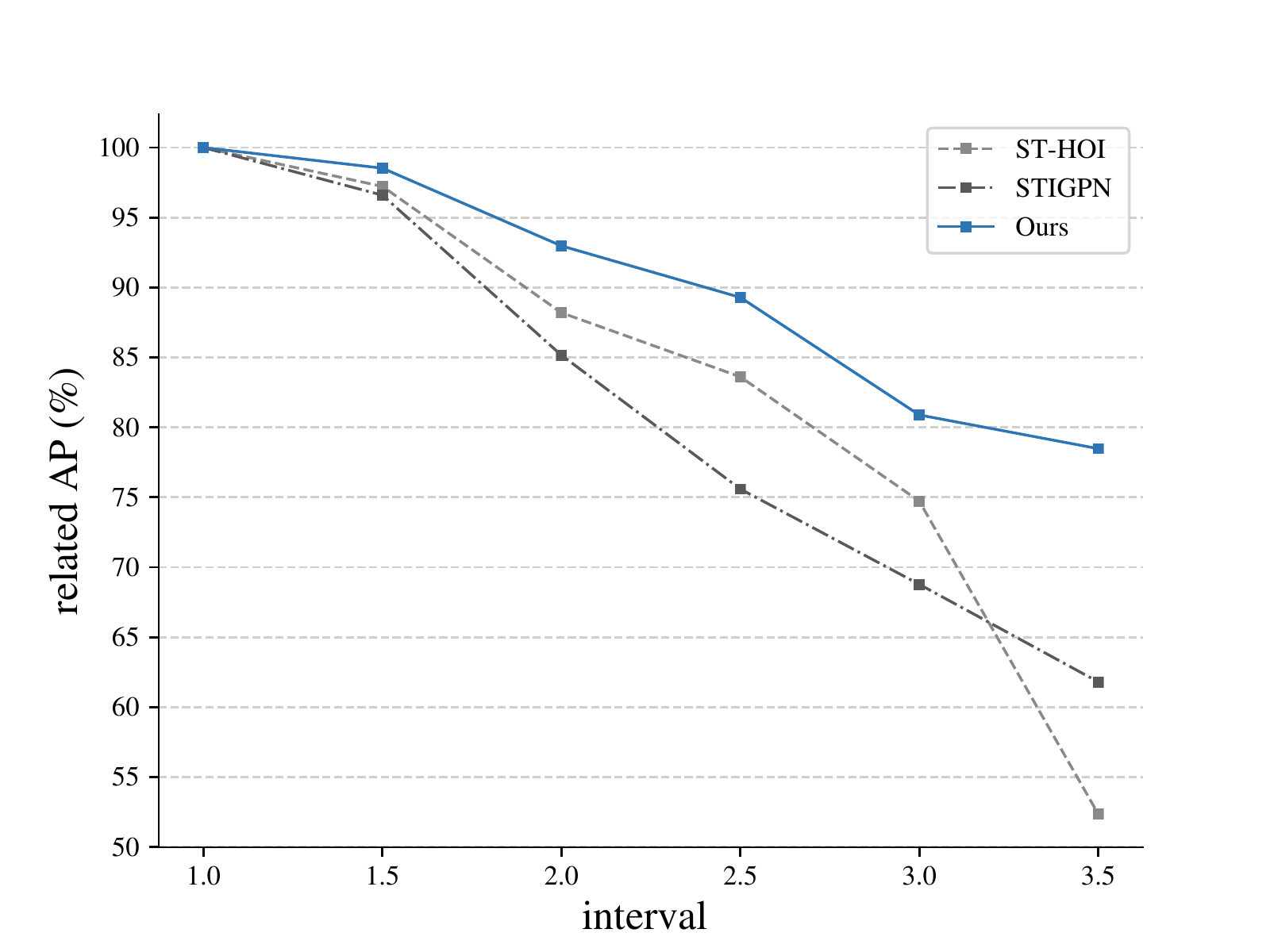}
    %\captionsetup{width=0.95\linewidth}
    %\captionsetup{size=small}
    %\caption{ (b)}
    \end{minipage}
    }
    \subfloat[\textbf{Similarity matrix}.
    \label{fig:sm}
    ]{
    \begin{minipage}[t]{0.32\textwidth}
    \centering
    \includegraphics[width=0.98\linewidth]{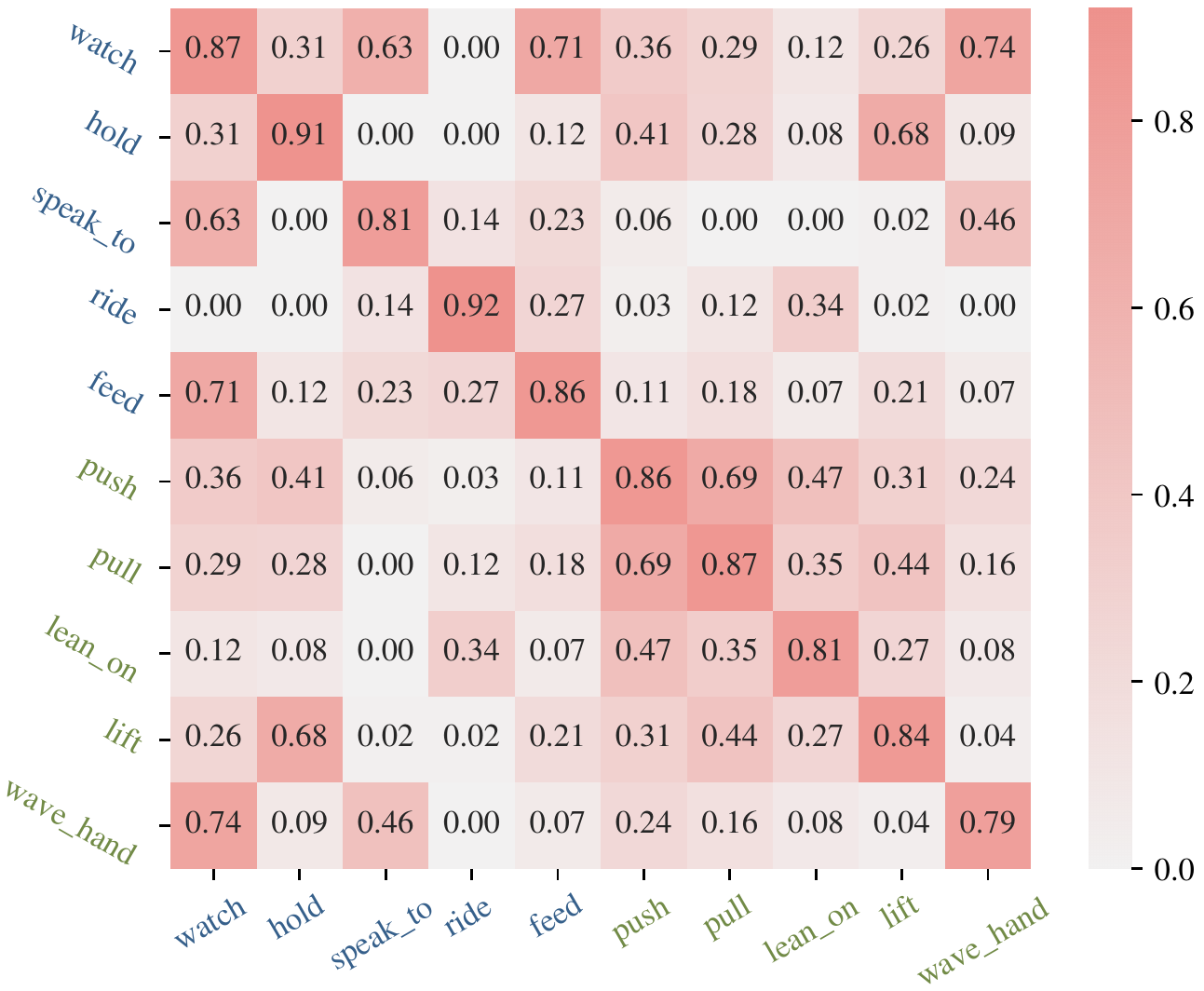}
    %\captionsetup{width=0.95\linewidth}
    %\captionsetup{size=small}
    %\caption{(c)}
    \end{minipage}
    }
    %\captionsetup{width=0.95\linewidth}
    %\captionsetup{size=small}
    \caption{The performance of CNN-based and Transformer-based methods under different scenarios.}
    %\vspace{-2em}
    \label{tran_cnn}
    \end{figure}

%\vspace{-7pt}
 We compare CNN-based and Transformer-based methods in terms of: 1) long-range dependency modeling, 2) robustness to time discontinuity and 3) contextual relation reasoning.\\
 {\bfseries Long-range dependency modeling.}
 We split HOI instances into bins of size 0.1 according to the normalized spatial distances, and report the APs of each bin. As shown in Figure~\ref{fig:dis}, our Transformer-based method outperforms existing CNN-based methods in all cases, which becomes increasingly evident as the spatial distance grows. It indicates that Transformer has better long-range dependency modeling capability compared to CNN-based methods that relay on limited receptive field. With this ability, Transformer can dynamically aggregate important information from global context. \\
 {\bfseries Robustness to time discontinuity.}
 we randomly sample one frame every $t$ seconds from the original video to generate a new video as inputs, and report the relative performance compared to the baseline (sampling 1 frame per second). As shown in Figure~\ref{fig:tem}, the performance of CNN-based methods drop dramatically in contrast to Transformer. The main reason is that the ROI features from different frames are likely to be inconsistent due to the discontinuity of temporal domain. For Transformer, it can be partly solved by learning a variable attention weights to selectively process different features of different frames.\\
 {\bfseries Contextual relations reasoning.}
 we randomly pick 5 static interaction types (represented in {\color{airforceblue}{blue}}) and 5 dynamic ones ({\color{palegreen}{green}}), each with over 10,000 images. Then, we calculate the average weights of self-attention in the last decoder layer on all pictures where two interactions are co-predicted. As shown in Figure~\ref{fig:sm}, Transformer can mine the interrelations among different HOI instances, \emph{e.g.}, \underline{watch} and \underline{feed}, which two are likely to co-occur, get a relatively high attention weights (0.71). 

%\vspace{-9pt}
\subsection{Analysis of Token abstraction and Linking}
%\vspace{-7pt}

{\bfseries Token abstraction.}
    Table.~\ref{tab:spatial} shows the influence of token abstraction, which is proposed to capture instance level representation. In comparison, CNN-based methods process cropped proposal features (Figure~\ref{first_img}(a)), which suffers from temporal inconsistencies and the lack of contextual information, leading to the worst performance. In terms of Transformer, performance is significantly improved (over $25\%$ improvement), but varies under different strategies. Interestingly, adding a simple token fusion module to ViT-like framework (ViT-like$^{\dag}$), \emph{i.e.}, fusing every 4 neighboring  patch tokens after each Transformer layer,  can achieve a $4\%$  relative mAP improvement. It implies that visual redundancy is an obstacle for Transformer to achieve better performance.  Moreover, our irregular-window-based (IR-win) token abstraction mechanism achieves the optimal performance. Nevertheless, when replacing all irregular windows in TUTOR with regular windows (R-win), the performance is unexpectedly surpassed by ViT-like$^{\dag}$. It indicates that regular windows can reduce the computational complexity, but cannot eliminate visual redundancy thoroughly.\\
{\bfseries Token linking.} 
    Table.~\ref{tab:temporal} shows the influence of token linking. Here, the inputs for all methods are identical, which are the instance tokens generated by token abstraction module. Although computing global attentions along temporal domain without token linking achieves a competitive performance in the detection of time-related interactions, its performance is relatively poor for detecting static HOIs. We conjecture that the instance tokens in different frames are semantically similar, which introduces redundant information to static interaction detection. Moreover, the mAP decreases severely when directly use the value of \texttt{Gumbel-Softmax}  as assignment weights, \emph{i.e.}, replace the $\hat{\mathbf{A}}$ in Eq.\ref{eq:temporal} with $\mathbf{A}$ in Eq.\ref{eq:softmax}. One possible reason is the redundancy arises within token representation due to the absence of zero value in $\mathbf{A}$. In contrast, one-hot assignment is sparse but cannot ensure an one-to-one assignment among frames, which can also cause ambiguity. In comparison, nms-one-hot assignment enforces every $T$ tokens (one per frame) to be linked, which minimizes the ambiguity and redundancy in token representation, thus achieving the optimal performance. We further investigate the effect of video length on these two assignment approaches. As shown in Figure~\ref{fig:temporal}, one-hot assignment surpasses nms-one-hot when a video is longer than 16 seconds, which is caused by the simple way of choosing exemplar frame, \emph{i.e.}, intuitively selecting the middle frame. when a video clip is long, the middle frame is semantically inconsistent with the frames that are temporally far away. We solve this problem by splitting a long video into uniform short clips and performing nms-one-hot assignment in each clip (nms-one-hot*), respectively, which yet introduces more computation.
    
%\vspace{-11pt}
\subsection{Analysis of Effectiveness and Efficiency}
%\vspace{-9pt}
\input{table/tokenization.tex}
\input{table/eff.tex}

{\bfseries Effectiveness.}
    We verify TUTOR's effectiveness of capturing spatial and temporal semantic by observing the performance of detecting static HOI and dynamic HOI when using a quite simple decoder. For former, we use a 1-layer Transformer decoder on patch tokens in ViT-like method and instance tokens in TUTOR, respectively. the mAP is reported only on static HOI detection. As the Table~\ref{tab:effect} shows, instance tokens generated by token abstraction can stupendously improve the performance by $70\%$, compared with patch tokens. It demonstrates that token abstraction mechanism can significantly extract highly-abstracted instance level semantic, which can be easily captured even the decoder is simple. For dynamic HOI detection, we use a 4-layer perception with RELU in between as decoder. Next, we perform a global average pooling on tubelet tokens and instance tokens, which are then fed to the simple decoder to predict the dynamic interactions, respectively. Interestingly, tubelet tokens generated by token linking boost performance by $4\times$, showing us the importance to reduce the temporal redundancy.\\
{\bfseries Efficiency.}
    Computing attention weights accounts for most of computational overhead in Transformer. Compared to the quadratic computational costs in global attention, we achieve a linear one. As shown in Table~\ref{tab:effic}, TUTOR achieves a $4\times$ speedup, greatly improving its usability in practical applications. Here, ``FPS'' is reported in terms of video, \emph{i.e.}, the number of videos being processed per second.

\input{table/ablation.tex}
\input{table/sota.tex}

%\vspace{-11pt}
\subsection{Ablation Study}
%\vspace{-9pt}

{\bfseries Token agglomeration.} 
    Token agglomeration is proposed to distill the token representation by selectively merging and projecting the semantically related tokens. We first intuitively try the k-mean~\cite{macqueen1967some}, an excellent classical clustering algorithm, but obtain a unexpectedly poor performance,  as shown in Table~\ref{tab:fusion}. The reason is two-fold:1) it is difficult to integrate the k-mean with the main network into an end-to-end pipeline; 2) it is hard to determine the value of K.  Then we replace irregular winodws in TUTOR with regular windows to merge every 4 neighboring tokens, an average pooling operation essentially, which reduces the feature redundancy to some extent. In comparison, irregular-window is more of an operation to selectively merge the semantically similar tokens to model highly abstracted features. It is worth emphasizing that gradually increase the dimension of agglomerated token is interestingly important, which achieves a gain of more than $6\%$ on mAP. We guess that the features are richer with increasing dimensions, as extensively adopted in CNNs.\\
{\bfseries Window size.} 
     Table~\ref{tab:w_s} varies the window size. The instances in an image could have variant sizes. Therefore, a small-sized window is hard to overlap different instances while a large-sized one could cause information mixture as unrelated tokens may be included. We find 7 to be optimal.\\
{\bfseries Small tricks.} We represent some small tricks for key module design in Table~\ref{tab:trick}. For \texttt{nms-one-hot} assignment, another commonly used strategy for merging assigned tokens is to concatenate them and then project them with a fully-connected layer. Compared with weighted-sum, it can slightly improve performance, but introduces more computation. For position encoding, we factorize the normally used 3D position encoding for spatiotemporal Transformer into a 2D spatial position encoding and a 1D temporal one, which two are added in token agglomeration and linking module, respectively. It is an experiential operation since spatial and temporal information are separately extracted.

%\vspace{-11pt}
\subsection{Comparison with State-of-the-art}
%\vspace{-9pt}
Unlike the popularity of image-based HOI detection, relatively less works investigate video-based one as a more practical yet challenging problem. Interestingly, the ability of image-based methods to detect dynamic HOI can be partly improved by replacing the original 2D backbone with a 3D one, but it weaken the ability of detecting static HOIs. With aforementioned strategies, our methods outperforms existing sota methods by a large margins. It is our belief that detecting HOI from video is more reasonable and practical since most interactions are time-related. Therefore, we hope our work will be useful for video-based human activity understanding research.

%\vspace{-13pt}
\section{Discussion \& Conclusion}
%\vspace{-11pt}

{\bfseries Limitation.}
\label{limitation}
    Our Transformer-based method suffers from the problem of overfitting when handling with small-scale datasets. In our experiments, we have to use the pertrained weights on VidHOI to initialize the model for CAD-120 (small scale), or performance would be severely degraded.\\
{\bfseries Broader impacts.}
    We know some applications that illegally analyze user behavior by video monitoring. Therefore, strict ethical review is essential to avoid our model being used for such applications.\\
{\bfseries Conclusion.}
In this paper, we represent TUTOR, a novel spatiotemporal Transformer for video-based HOI detection, which structurizes a video into a few tubelet tokens. To generate compact and expressive tubelet tokens, we propose a token abstraction scheme built on selective attention and token agglomeration, along with token linking strategy to link semantically-related tokens across frames. Our methods outperforms existing works by large margins. Going further, visual redundancy is one of the biggest obstacles for vision Transformer to achieve the same excellent performance as language Transformer, and we will devote more exploration on this in the future works.

{\small
	\bibliographystyle{plain}
	\bibliography{egbib}
}

\newpage
\appendix
\section{Appendix}
\begin{figure}[ht]
%\vspace{-0.3cm}
\centering
\includegraphics[width=\linewidth]{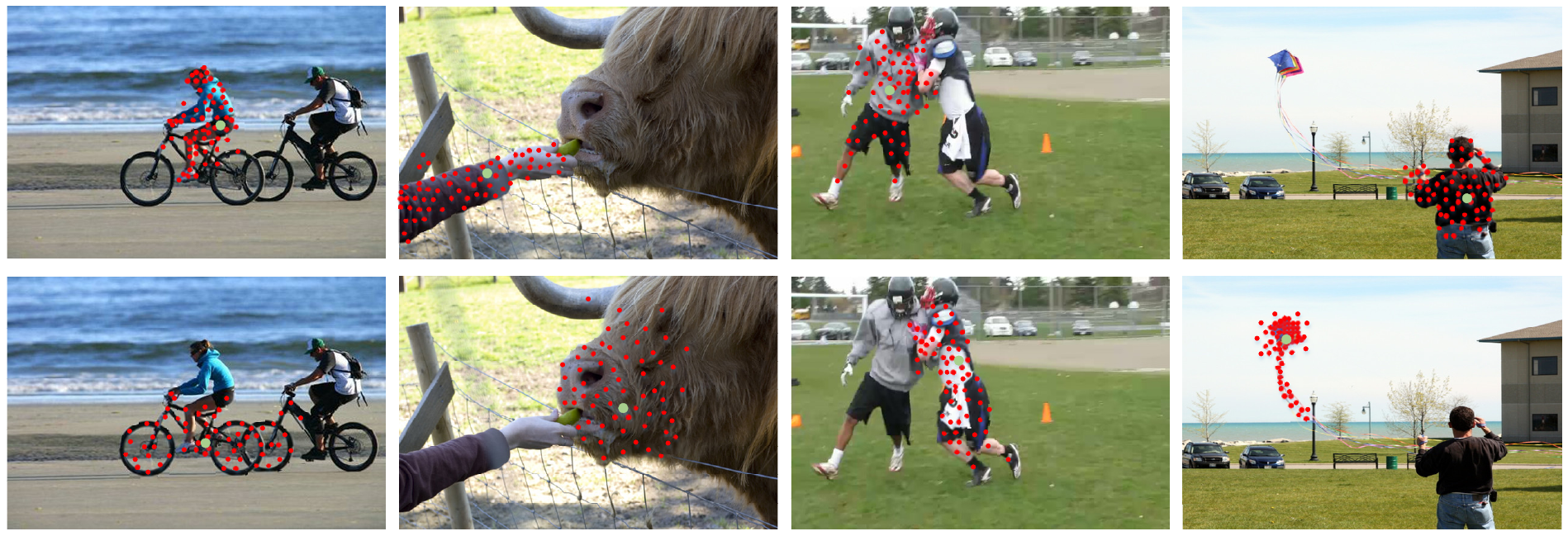}
%\vspace{-1.2em}
\caption{\textbf{Visualization of irregular windows.} Each {\color{palegreen}{green}} point is the location of an instance token obtained by token token abstraction, and the {\color{red}{red}} points are the locations of the tokens involved in agglomeration. There are totally $4^3=64$ sampling points for each instance token, since every 4 tokens are agglomerated in a token agglomeration layer and 3 layers are employed. As the input of TUTOR are feature maps ouputted by the backbone, the locations in the original image are calculated via bilinear interpolation.} 
\label{irw}
\end{figure}
\begin{figure}[ht]
%\vspace{-0.3cm}
\centering
\includegraphics[width=\linewidth]{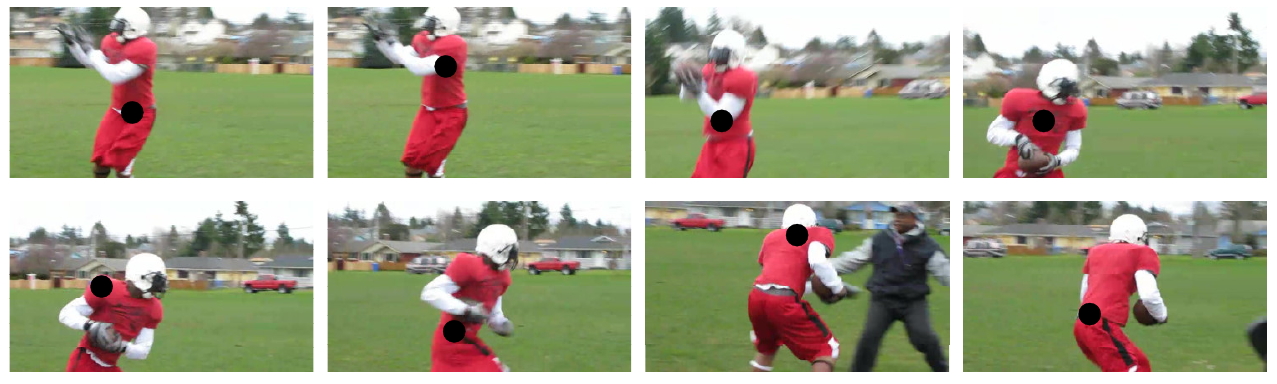}
%\vspace{-1.2em}

\caption{\textbf{Visualization of token linking.} The {color{black}{black}} points in different frames are the locations of the instance tokens involved in token linking, \emph{i.e.}, the instance tokens being linked in the same tubelet token.} 
%\vspace{-2em}
\label{tl}
\end{figure}

\end{document}

%% file: table/tokenization.tex
\begin{table*}[t]
%\vspace{-.1em}
    \centering
    \subfloat[\textbf{Token abstraction}.
    \label{tab:spatial}
    ]{
    \centering
    \begin{minipage}{0.28\linewidth}{\begin{center}
    \tablestyle{4pt}{1.3}
    \begin{tabular}{y{35}x{24}x{24}}
    method & S & T \\
    \shline
    Proposal & 22.84 & 16.34 \\
    ViT-like & 27.64 & 17.30 \\
    $\text{ViT-like}^{\dag}$ & 28.45 & 18.64 \\
    R-win  & 28.17 & 18.24   \\
    IR-win & \baseline{\textbf{32.21}} & \baseline{\textbf{21.28}}\\
    \end{tabular}
    \end{center}
    }\end{minipage}
    }
    \hspace{1em}
    \subfloat[\textbf{Token linking}.
    \label{tab:temporal}
    ]{
    \centering
    \begin{minipage}{0.28\linewidth}{\begin{center}
    \tablestyle{1pt}{1.3}
    \begin{tabular}{y{65}x{24}x{24}}
    method & S & T \\
    \shline
    global & 30.07 & 19.58 \\
    gumbel-softmax & 28.81 & 18.11 \\
    one-hot & 30.64 & 19.27 \\
    nms-one-hot & \baseline{\textbf{32.21}} & \baseline{\textbf{21.28}}\\
    \multicolumn{3}{c}{~}\\
    \end{tabular}
    \end{center}
    }\end{minipage}
    }
    \hspace{1em}
    \subfloat[\textbf{Tokenization vs. video length}.
    \label{fig:temporal}
    ]{
    \centering
    \begin{minipage}{0.31\linewidth}{\begin{center}
    \includegraphics[width=0.99\linewidth, height=2.7cm]{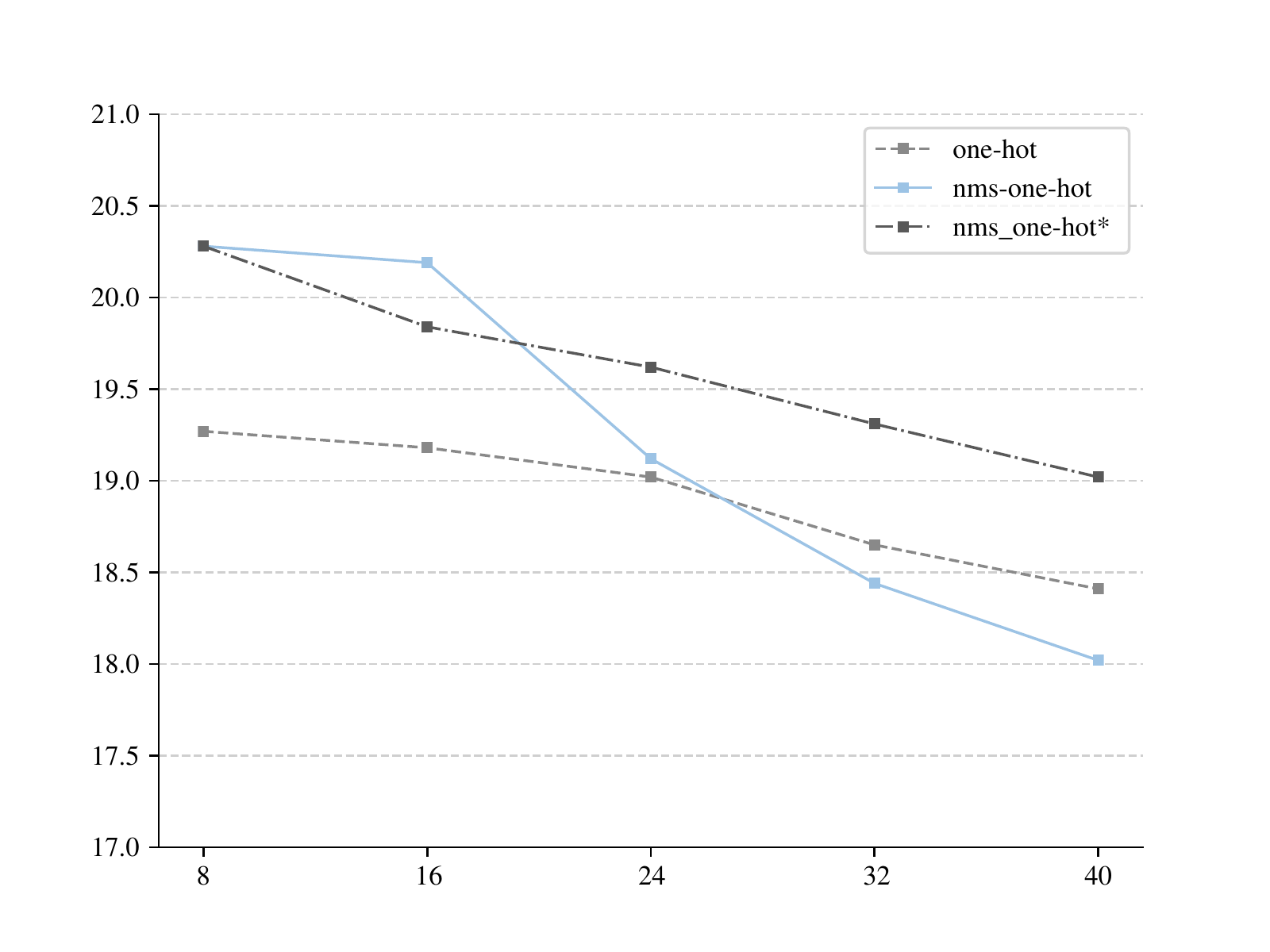}
    \end{center}
    }\end{minipage}
    }
    %\vspace{-0.3em}
    \caption{\textbf{Analysis on token abstraction and linking.} We report mAP on detecting dynamic temporal-related (T) and static spatial-related (S) HOI, respectively. global$^{\dag}$ in (a) denotes that a regular-window-based token fusing is performed after each Transformer layer.  Nms-one-hot$^*$ in (c) means to split a long video  into several uniform short clips. Default settings are marked in \colorbox{baselinecolor}{gray}.}
   %\vspace{-0.7em}
\end{table*}

%% file: table/eff.tex
\begin{table*}[t]
%\vspace{-.1em}
    \centering
    \subfloat[\textbf{Effectiveness}.
    \label{tab:effect}
    ]{
    \centering
    \begin{minipage}{0.31\linewidth}{\begin{center}
    \tablestyle{1pt}{1.3}
    \begin{tabular}{x{40}|x{55}x{24}}
     & case & mAP\\
    \shline
    \multirow{2}{*}[0em]{spatial} & {w/ } \texttt{TA} & \baseline{\textbf{16.42}} \\
     & {w/o } \texttt{TA}  & 9.67\\
     \shline
    \multirow{2}{*}[0em]{temporal} & {w/ } \texttt{TL} & \baseline{\textbf{8.28}} \\
     & {w/o } \texttt{TL} &  2.30\\
    \end{tabular}
    \end{center}
    }\end{minipage}
    }
    \hspace{2em}
    \subfloat[\textbf{Efficiency}.
    \label{tab:effic}
    ]{
    \centering
    \begin{minipage}{0.58\linewidth}{\begin{center}
    \tablestyle{1pt}{1.3}
    \begin{tabular}{y{55}|x{35}x{30}x{35}x{24}x{30}}
     case & params & mAP & TFLOPs & FPS &speedup\\
    \shline
    global & 243M & 23.51 & 0.81 & 0.5 & - \\
    {w/ } \texttt{TA} & 104M & 25.63 & 0.42 &  1.2 & $2 \times$ \\
    {w/ } \texttt{TL} & 187M & 24.28 & 0.76 & 0.8 & - \\
    {w/ } \texttt{(TA+TL)} & \baseline{\textbf{82M}} & \baseline{\textbf{26.84}} & \baseline{\textbf{0.25}} & \baseline{\textbf{2.0}} & \baseline{$\mathbf{4 \times}$} \\
    \end{tabular}
    \end{center}
    }\end{minipage}
    }
   %\vspace{-0.3em}
   \caption{\textbf{Analysis on effectiveness and efficiency.} \texttt{TA} is short for token agglomeration and \texttt{TL} for token linking. We use clips of size $8 \times 384 \times 384$, with frames sampled at a rate of 1/32.}
   %\vspace{-1.9em}
\end{table*}

%% file: table/ablation.tex
\begin{table*}[t]
    \centering
    \subfloat[\textbf{Token agglomeration}.
    \label{tab:fusion}
    ]{
    \centering
    \begin{minipage}{0.28\linewidth}{\begin{center}
    \tablestyle{4pt}{1.2}
    \begin{tabular}{y{40}x{20}x{20}}
    method & S & T \\
    \shline
    k-mean & 26.71 & 17.56 \\
    r-win & 28.82 & 18.95 \\
    ir-win-C & 29.34 & 19.43 \\
    ir-win-2C & \baseline{\textbf{32.21}} & \baseline{\textbf{21.28}}\\
    \end{tabular}
    \end{center}
    }\end{minipage}
    }
   %\hspace{0.1em}
    \subfloat[\textbf{Window size}.
    \label{tab:w_s}
    ]{
    \centering
    \begin{minipage}{0.28\linewidth}{\begin{center}
    \tablestyle{4pt}{1.2}
    \begin{tabular}{x{20}x{20}x{20}}
    size & S & T \\
    \shline
    3 & 24.80 & 17.92 \\
    5 & 30.69 & 19.75 \\
    7 & \baseline{\textbf{32.21}} & \baseline{\textbf{21.28}} \\
    11 & 30.28 & 19.71 
    \end{tabular}
    \end{center}
    }\end{minipage}
    }
    %\hspace{0.1em}
    \subfloat[\textbf{Small tricks}.
    \label{tab:trick}
    ]{
    \centering
    \begin{minipage}{0.31\linewidth}{\begin{center}
    \tablestyle{1pt}{1.2}
    \begin{tabular}{y{49}|x{35}x{24}x{24}}
    component & trick & S & T \\
    \shline
    \multirow{2}{*}[0em]{nms-one-hot} & concat. & \textbf{32.85} & \textbf{21.44} \\
     & w-sum  & \baseline{32.21} & \baseline{21.28}\\
     \shline
    \multirow{2}{*}[0em]{p-encoding} & 3D & 31.62 & 20.59 \\
     & (2+1)D &  \baseline{\textbf{32.21}} & \baseline{\textbf{21.28}}\\
    \end{tabular}
    \end{center}
    }\end{minipage}
    }
   %\vspace{-0.3em}
   \caption{\textbf{Ablation study}. In (a), ``$n$C'' is the dimension of agglomerated token. in (c), ``p-encoding' is short for position encoding and ``w-sum'' for weighted-sum.}
   %\vspace{-0.8em}
\end{table*}

%% file: table/sota.tex
\begin{table*}[t]
%\vspace{-.1em}
    \centering

    \tablestyle{1.3pt}{1.3}
    \begin{tabular}{y{80}|x{50}x{24}|x{30}x{30}x{34}x{24}x{30}|x{54}}
    
     \multirow{2}{*}[0em]{method} & \multirow{2}{*}[0em]{backbone} & \multirow{2}{*}[0em]{P} & \multicolumn{5}{c|}{\scriptsize VidHOI} & \scriptsize CAD-120\\
    & & & \scriptsize Full & \scriptsize Rare & \scriptsize NoneRare & \scriptsize S & \scriptsize T & \scriptsize sub-activity($\%$)\\
    \shline
     \gc{\scriptsize \textit{CNN-based methods}} & & & & & & & &\\
     \scriptsize PMF~\cite{wan2019pose} w/ SlowFast & \scriptsize SlowFast~\cite{feichtenhofer2019slowfast} & $\checkmark$  & 16.31 & 14.28  & 23.86 & 21.77 & 8.42 & - \\
     \scriptsize GPNN~\cite{qi2018learning}  & \scriptsize ResNet-101 &  & 18.47 & 16.41  & 24.50 & 26.41 & 16.06 & 88.9 \\
     \scriptsize STIGPN~\cite{wang2021spatio}  & \scriptsize ResNet-50 &  & 19.39 & 18.22  & 28.13 & 26.58 & 18.46 & 91.9 \\
     \scriptsize ST-HOI~\cite{chiou2021st}  & \scriptsize SlowFast & $\checkmark$  & 17.60 & 17.30  & 27.20 & 25.00 & 14.40 & - \\
     \shline
     \gc{\scriptsize \textit{Transformer-based methods}} & & & & & & & &\\
     \scriptsize HOTR*~\cite{hotr}  & \scriptsize ResNet-50 &  & 21.14 & 19.83  & 30.75 & 28.36 & 9.81 & - \\
     \scriptsize QPIC*~\cite{hotr}  & \scriptsize ResNet-50 &  & 21.40 & 20.56  & 32.90 & 28.87 & 9.74 & - \\
     \scriptsize HOTR w/ SlowFast  & \scriptsize SlowFast &  & 22.84 & 21.15  & 32.86 & 27.12 & 13.29 & 90.7 \\
     \scriptsize QPIC w/ SlowFast  & \scriptsize SlowFast &  & 22.92 & 21.64  & 33.43 & 28.41 & 13.47 & 91.3 \\
     \scriptsize TimeSformer~\cite{bertasius2021space}  w/ decoder  & \scriptsize TimeSformer &  & 23.17 & 21.79  & 34.57 & 27.84 & 18.90 & 92.5 \\
     \scriptsize ours  & \scriptsize ResNet-50 &  & \baseline{\textbf{26.92}} & \baseline{\textbf{23.49}}  & \baseline{\textbf{37.12}} & \baseline{\textbf{32.21}} & \baseline{\textbf{21.28}} & \baseline{\textbf{94.7}} \\
    \end{tabular}
    
   %\vspace{-0.3em}
   \caption{\textbf{comparison with state-of-the-art.} P means human poses, $^*$ denotes image-based method.}
   %\vspace{-2em}
\end{table*}

%% file: neurips_2022.bbl
\begin{thebibliography}{10}

\bibitem{bertasius2021space}
Gedas Bertasius, Heng Wang, and Lorenzo Torresani.
\newblock Is space-time attention all you need for video understanding.
\newblock In {\em ICML}, 2021.

\bibitem{chao2018learning}
Yu-Wei Chao, Yunfan Liu, Xieyang Liu, Huayi Zeng, and Jia Deng.
\newblock Learning to detect human-object interactions.
\newblock In {\em WACV}, 2018.

\bibitem{chen2021reformulating}
Mingfei Chen, Yue Liao, Si~Liu, Zhiyuan Chen, Fei Wang, and Chen Qian.
\newblock Reformulating hoi detection as adaptive set prediction.
\newblock In {\em CVPR}, 2021.

\bibitem{chiou2021st}
Meng-Jiun Chiou, Chun-Yu Liao, Li-Wei Wang, Roger Zimmermann, and Jiashi Feng.
\newblock St-hoi: A spatial-temporal baseline for human-object interaction
  detection in videos.
\newblock In {\em ICDAR}, pages 9--17, 2021.

\bibitem{dosovitskiy2020image}
Alexey Dosovitskiy, Lucas Beyer, Alexander Kolesnikov, Dirk Weissenborn,
  Xiaohua Zhai, Thomas Unterthiner, Mostafa Dehghani, Matthias Minderer, Georg
  Heigold, Sylvain Gelly, et~al.
\newblock An image is worth 16x16 words: Transformers for image recognition at
  scale.
\newblock In {\em ICLR}, 2021.

\bibitem{esser2021taming}
Patrick Esser, Robin Rombach, and Bjorn Ommer.
\newblock Taming transformers for high-resolution image synthesis.
\newblock In {\em CVPR}, pages 12873--12883, 2021.

\bibitem{feichtenhofer2019slowfast}
Christoph Feichtenhofer, Haoqi Fan, Jitendra Malik, and Kaiming He.
\newblock Slowfast networks for video recognition.
\newblock In {\em CVPR}, 2019.

\bibitem{drg}
Chen Gao, Jiarui Xu, Yuliang Zou, and Jia-Bin Huang.
\newblock Drg: Dual relation graph for human-object interaction detection.
\newblock In {\em ECCV}, 2020.

\bibitem{ican}
Chen Gao, Yuliang Zou, and Jia-Bin Huang.
\newblock ican: Instance-centric attention network for human-object interaction
  detection.
\newblock In {\em BMVC}, 2018.

\bibitem{garcia2020knowledge}
Noa Garcia and Yuta Nakashima.
\newblock Knowledge-based video question answering with unsupervised scene
  descriptions.
\newblock In {\em ECCV}, pages 581--598, 2020.

\bibitem{interactions}
Georgia Gkioxari, Ross Girshick, Piotr Doll{\'a}r, and Kaiming He.
\newblock Detecting and recognizing human-object interactions.
\newblock In {\em CVPR}, 2018.

\bibitem{no-frills}
Tanmay Gupta, Alexander Schwing, and Derek Hoiem.
\newblock No-frills human-object interaction detection: Factorization, layout
  encodings, and training techniques.
\newblock In {\em ICCV}, 2019.

\bibitem{he2016deep}
Kaiming He, Xiangyu Zhang, Shaoqing Ren, and Jian Sun.
\newblock Deep residual learning for image recognition.
\newblock In {\em CVPR}, 2016.

\bibitem{HIC}
MD.~Zakir Hossain, Ferdous Sohel, Mohd~Fairuz Shiratuddin, and Hamid Laga.
\newblock A comprehensive survey of deep learning for image captioning.
\newblock {\em CsUR}, 2019.

\bibitem{zhi_eccv2020}
Zhi Hou, Xiaojiang Peng, Yu~Qiao, and Dacheng Tao.
\newblock Visual compositional learning for human-object interaction detection.
\newblock In {\em ECCV}, 2020.

\bibitem{jang2016categorical}
Eric Jang, Shixiang Gu, and Ben Poole.
\newblock Categorical reparameterization with gumbel-softmax.
\newblock In {\em ICLR}, 2017.

\bibitem{ji2021detecting}
Jingwei Ji, Rishi Desai, and Juan~Carlos Niebles.
\newblock Detecting human-object relationships in videos.
\newblock In {\em ICCV}, pages 8106--8116, 2021.

\bibitem{uniondet}
Bumsoo Kim, Taeho Choi, Jaewoo Kang, and Hyunwoo~J. Kim.
\newblock Union{D}et: Union-level detector towards real-time human-object
  interaction detection.
\newblock In {\em ECCV}, 2020.

\bibitem{hotr}
Bumsoo Kim, Junhyun Lee, Jaewoo Kang, Eun-Sol Kim, and Hyunwoo~J Kim.
\newblock Hotr: End-to-end human-object interaction detection with
  transformers.
\newblock In {\em CVPR}, 2021.

\bibitem{kim2020detecting}
Dong-Jin Kim, Xiao Sun, Jinsoo Choi, Stephen Lin, and In~So Kweon.
\newblock Detecting human-object interactions with action co-occurrence priors.
\newblock In {\em ECCV}, 2020.

\bibitem{koppula2013learning}
Hema~Swetha Koppula, Rudhir Gupta, and Ashutosh Saxena.
\newblock Learning human activities and object affordances from rgb-d videos.
\newblock {\em IJRR}, pages 951--970, 2013.

\bibitem{kuhn1955hungarian}
Harold~W Kuhn.
\newblock The hungarian method for the assignment problem.
\newblock {\em Naval Research Logistics Quarterly}, 1955.

\bibitem{li2020detailed}
Yong-Lu Li, Xinpeng Liu, Han Lu, Shiyi Wang, Junqi Liu, Jiefeng Li, and Cewu
  Lu.
\newblock Detailed 2d-3d joint representation for human-object interaction.
\newblock In {\em CVPR}, 2020.

\bibitem{li2019transferable}
Yong-Lu Li, Siyuan Zhou, Xijie Huang, Liang Xu, Ze~Ma, Hao-Shu Fang, Yanfeng
  Wang, and Cewu Lu.
\newblock Transferable interactiveness knowledge for human-object interaction
  detection.
\newblock In {\em CVPR}, 2019.

\bibitem{liang2022vrt}
Jingyun Liang, Jiezhang Cao, Yuchen Fan, Kai Zhang, Rakesh Ranjan, Yawei Li,
  Radu Timofte, and Luc Van~Gool.
\newblock Vrt: A video restoration transformer.
\newblock {\em arXiv preprint arXiv:2201.12288}, 2022.

\bibitem{lin2017feature}
Tsung-Yi Lin, Piotr Doll{\'a}r, Ross Girshick, Kaiming He, Bharath Hariharan,
  and Serge Belongie.
\newblock Feature pyramid networks for object detection.
\newblock In {\em CVPR}, 2017.

\bibitem{lin2020action}
Xue Lin, Qi~Zou, and Xixia Xu.
\newblock Action-guided attention mining and relation reasoning network for
  human-object interaction detection.
\newblock In {\em IJCAI}, 2020.

\bibitem{liu2021end}
Xiaolong Liu, Qimeng Wang, Yao Hu, Xu~Tang, Song Bai, and Xiang Bai.
\newblock End-to-end temporal action detection with transformer.
\newblock {\em arXiv preprint arXiv:2106.10271}, 2021.

\bibitem{liu2020amplifying}
Yang Liu, Qingchao Chen, and Andrew Zisserman.
\newblock Amplifying key cues for human-object-interaction detection.
\newblock In {\em ECCV}, 2020.

\bibitem{swin}
Ze~Liu, Yutong Lin, Yue Cao, Han Hu, Yixuan Wei, Zheng Zhang, Stephen Lin, and
  Baining Guo.
\newblock Swin transformer: Hierarchical vision transformer using shifted
  windows.
\newblock In {\em ICCV}, 2021.

\bibitem{adamw}
Ilya Loshchilov and Frank Hutter.
\newblock Decoupled weight decay regularization.
\newblock In {\em ICLR}, 2017.

\bibitem{macqueen1967some}
James MacQueen et~al.
\newblock Some methods for classification and analysis of multivariate
  observations.
\newblock In {\em Proceedings of the fifth Berkeley symposium on mathematical
  statistics and probability}, 1967.

\bibitem{nagarajan2019grounded}
Tushar Nagarajan, Christoph Feichtenhofer, and Kristen Grauman.
\newblock Grounded human-object interaction hotspots from video.
\newblock In {\em ICCV}, pages 8688--8697, 2019.

\bibitem{nawhal2020generating}
Megha Nawhal, Mengyao Zhai, Andreas Lehrmann, Leonid Sigal, and Greg Mori.
\newblock Generating videos of zero-shot compositions of actions and objects.
\newblock In {\em ECCV}, pages 382--401, 2020.

\bibitem{qi2018learning}
Siyuan Qi, Wenguan Wang, Baoxiong Jia, Jianbing Shen, and Song-Chun Zhu.
\newblock Learning human-object interactions by graph parsing neural networks.
\newblock In {\em ECCV}, 2018.

\bibitem{sunkesula2020lighten}
Sai Praneeth~Reddy Sunkesula, Rishabh Dabral, and Ganesh Ramakrishnan.
\newblock Lighten: Learning interactions with graph and hierarchical temporal
  networks for hoi in videos.
\newblock In {\em ACMMM}, pages 691--699, 2020.

\bibitem{qpic}
Masato Tamura, Hiroki Ohashi, and Tomoaki Yoshinaga.
\newblock Qpic: Query-based pairwise human-object interaction detection with
  image-wide contextual information.
\newblock In {\em CVPR}, 2021.

\bibitem{ulutan2020vsgnet}
Oytun Ulutan, ASM Iftekhar, and Bangalore~S Manjunath.
\newblock Vsgnet: Spatial attention network for detecting human object
  interactions using graph convolutions.
\newblock In {\em CVPR}, 2020.

\bibitem{attention}
Ashish Vaswani, Noam Shazeer, Niki Parmar, Jakob Uszkoreit, Llion Jones,
  Aidan~N Gomez, {\L}ukasz Kaiser, and Illia Polosukhin.
\newblock Attention is all you need.
\newblock In {\em NIPS}, 2017.

\bibitem{wan2019pose}
Bo~Wan, Desen Zhou, Yongfei Liu, Rongjie Li, and Xuming He.
\newblock Pose-aware multi-level feature network for human object interaction
  detection.
\newblock In {\em ICCV}, 2019.

\bibitem{wang2020contextual}
Hai Wang, Wei-shi Zheng, and Ling Yingbiao.
\newblock Contextual heterogeneous graph network for human-object interaction
  detection.
\newblock In {\em ECCV}, 2020.

\bibitem{wang2021spatio}
Ning Wang, Guangming Zhu, Liang Zhang, Peiyi Shen, Hongsheng Li, and Cong Hua.
\newblock Spatio-temporal interaction graph parsing networks for human-object
  interaction recognition.
\newblock In {\em ACMMM}, pages 4985--4993, 2021.

\bibitem{wang2019deep}
Tiancai Wang, Rao~Muhammad Anwer, Muhammad~Haris Khan, Fahad~Shahbaz Khan,
  Yanwei Pang, Ling Shao, and Jorma Laaksonen.
\newblock Deep contextual attention for human-object interaction detection.
\newblock In {\em ICCV}, 2019.

\bibitem{wang2021end}
Yuqing Wang, Zhaoliang Xu, Xinlong Wang, Chunhua Shen, Baoshan Cheng, Hao Shen,
  and Huaxia Xia.
\newblock End-to-end video instance segmentation with transformers.
\newblock In {\em CVPR}, pages 8741--8750, 2021.

\bibitem{xu2019interact}
Bingjie Xu, Junnan Li, Yongkang Wong, Qi~Zhao, and Mohan~S Kankanhalli.
\newblock Interact as you intend: Intention-driven human-object interaction
  detection.
\newblock {\em TMM}, 2019.

\bibitem{yang2020graph}
Dongming Yang and Yuexian Zou.
\newblock A graph-based interactive reasoning for human-object interaction
  detection.
\newblock {\em IJCAI}, 2020.

\bibitem{zhang2021mining}
Aixi Zhang, Yue Liao, Si~Liu, Miao Lu, Yongliang Wang, Chen Gao, and Xiaobo Li.
\newblock Mining the benefits of two-stage and one-stage hoi detection.
\newblock {\em NIPs}, 34, 2021.

\bibitem{zhang2022actionformer}
Chenlin Zhang, Jianxin Wu, and Yin Li.
\newblock Actionformer: Localizing moments of actions with transformers.
\newblock {\em arXiv preprint arXiv:2202.07925}, 2022.

\bibitem{zhong2021polysemy}
Xubin Zhong, Changxing Ding, Xian Qu, and Dacheng Tao.
\newblock Polysemy deciphering network for robust human--object interaction
  detection.
\newblock In {\em ICCV}, 2021.

\bibitem{zhou2019relation}
Penghao Zhou and Mingmin Chi.
\newblock Relation parsing neural network for human-object interaction
  detection.
\newblock In {\em ICCV}, 2019.

\bibitem{zhou2020cascaded}
Tianfei Zhou, Wenguan Wang, Siyuan Qi, Haibin Ling, and Jianbing Shen.
\newblock Cascaded human-object interaction recognition.
\newblock In {\em CVPR}, 2020.

\bibitem{deformableDETR}
Xizhou Zhu, Weijie Su, Lewei Lu, Bin Li, Xiaogang Wang, and Jifeng Dai.
\newblock Deformable detr: Deformable transformers for end-to-end object
  detection.
\newblock In {\em ICLR}, 2020.

\bibitem{endd}
Cheng Zou, Bohan Wang, Yue Hu, Junqi Liu, Qian Wu, Yu~Zhao, Boxun Li, Chenguang
  Zhang, Chi Zhang, and Yichen Wei.
\newblock End-to-end human object interaction detection with hoi transformer.
\newblock In {\em CVPR}, 2021.

\end{thebibliography}
